\documentclass[acmsmall]{acmart}
\AtBeginDocument{%
  }

\setcopyright{acmlicensed}
\copyrightyear{2025}
\acmYear{2025}
\acmDOI{XXXXXXX.XXXXXXX}

\acmJournal{TKDD}

\usepackage{tabulary}
\usepackage{graphicx}%
\usepackage{multirow}%
\usepackage{mathrsfs}%
\usepackage[title]{appendix}%
\usepackage{xcolor}%
\usepackage{textcomp}%
\usepackage{manyfoot}%
\usepackage{booktabs}%
\usepackage{algorithm}%
\usepackage{algorithmicx}%
\usepackage{algpseudocode}%
\usepackage{listings}%
\usepackage{hyperref}
\usepackage{cleveref}
\usepackage{threeparttable}
\usepackage{array}
\usepackage{enumitem}

\usepackage{arydshln} 

\begin{document}

\title{BiDepth: A Bidirectional-Depth Neural Network for Spatio-Temporal Prediction}
\author{Sina Ehsani}

\orcid{0000-0002-6009-7612}
\affiliation{%
  \institution{University of Arizona}
  \city{Tucson}
  \state{Arizona}
  \country{USA}
  \postcode{85721}
}
\email{sinaehsani@arizona.edu}

\author{Fenglian Pan}
\orcid{0009-0001-9599-636X}
\affiliation{%
  \institution{University of Arizona}
  \city{Tucson}
  \state{Arizona}
  \country{USA}
  \postcode{85721}}
\email{fenglianpan@arizona.edu}

\author{Qingpei Hu}
\orcid{0000-0003-3702-974x}
\affiliation{%
  \institution{Chinese Academy of Sciences}
  \city{Beijing}
  \country{China}
  \postcode{100190}}
\email{qingpeihu@amss.ac.cn}

\author{Jian Liu}
\orcid{0000-0003-0268-2941}
\affiliation{%
  \institution{University of Arizona}
  \city{Tucson}
  \state{Arizona}
  \country{USA}
  \postcode{85721}}
\email{jianliu@arizona.edu}

\renewcommand{\shortauthors}{Ehsani et al.}

\begin{abstract}
 Accurate spatial-temporal (ST) prediction for dynamic systems, such as urban mobility and weather patterns, is crucial but hindered by complex ST correlations and the challenge of concurrently modeling long-term trends with short-term fluctuations. Existing methods often falter in these areas. This paper proposes the BiDepth Multimodal Neural Network (BDMNN), which integrates two key innovations: 1) a bidirectional depth modulation mechanism that dynamically adjusts network depth to comprehensively capture both long-term seasonality and immediate short-term events; and 2) a novel convolutional self-attention cell (CSAC). Critically, unlike many attention mechanisms that can lose spatial acuity, our CSAC is specifically designed to preserve crucial spatial relationships throughout the network, akin to standard convolutional layers, while simultaneously capturing temporal dependencies. Evaluated on real-world urban traffic and precipitation datasets, BDMNN demonstrates significant accuracy improvements, achieving a 12\% Mean Squared Error (MSE) reduction in urban traffic prediction and a 15\% improvement in precipitation forecasting over leading deep learning benchmarks like ConvLSTM, using comparable computational resources. These advancements offer robust ST forecasting for smart city management, disaster prevention, and resource optimization.
\end{abstract}




\maketitle

\section{Introduction}
\label{sec:introduction}

Spatio-temporal (ST) data, characterized by observations that vary across both spatial and temporal dimensions, are increasingly prevalent and critical in diverse domains such as climate science \citep{aburas2019spatio}, urban planning \citep{song2019spatio, xia2025penalized}, transportation systems \citep{ling2023sthan, ehsani2024predicting}, and environmental monitoring \citep{amato2020novel}. In these fields, accurate forecasting is paramount, enabling proactive resource allocation in transportation systems, enhanced urban planning, and timely disaster mitigation based on environmental patterns. These datasets encapsulate complex ST dynamics, offering invaluable insights into evolving patterns and trends. A prime example is the New York City (NYC) Taxi and Limousine Commission Trip Record Data (``TLC data'' hereafter - refer to \Cref{sec:tlc_data} for details), providing rich information on taxi trips \citep{taxi2017tlc}, the analysis of which is crucial for understanding urban mobility and optimizing transportation \citep{zhou2019spatiotemporal}. The central challenge, therefore, is to develop models that can autonomously discover these intricate, multi-scale dependencies from raw data to generate accurate forecasts.

However, the inherent complexity of ST data presents significant forecasting challenges \citep{yuan2021survey}. Firstly, models must effectively capture intricate \textbf{ST correlations} \citep{li2017diffusion}. This involves understanding not only how phenomena at nearby locations are related (spatial correlations) but also how current states depend on past states (temporal correlations), and critically, how these spatial patterns evolve over time. Many contemporary deep learning approaches, especially those adapting attention mechanisms primarily designed for sequential data, can struggle to preserve and fully leverage vital spatial information throughout the network. Secondly, ST data are often characterized by \textbf{non-stationary patterns}, exhibiting a mixture of long-term seasonalities or trends and abrupt, short-term fluctuations \citep{lai2018modeling}. Existing models frequently face a trade-off, either over-smoothing predictions by focusing on long-term dynamics or becoming too sensitive to noise by overly emphasizing short-term variations, thus failing to achieve an optimal balance. Thirdly, the often \textbf{high dimensionality} of ST data, arising from numerous spatial locations and extensive time series, adds another layer of computational and modeling complexity \citep{geng2019spatiotemporal}.

These persistent limitations in comprehensively addressing ST correlations, non-stationary dynamics, and high dimensionality highlight the need for novel architectures. This paper introduces the \textbf{BiDepth Multimodal Neural Network (BDMNN)}, a deep learning framework specifically engineered to tackle these multifaceted challenges in ST forecasting. BDMNN's innovation is rooted in two core components: \textit{A)} \textbf{BiDepth mechanism} featuring a unique bidirectional depth modulation strategy. This allows the network to dynamically adjust its analytical depth for different temporal segments of an input sequence, enabling a nuanced balance between capturing long-range historical dependencies and responding effectively to recent, transient patterns. \textit{B)} A novel \textbf{Convolutional Self-Attention Cell (CSAC)}. Critically, unlike many attention mechanisms that can disrupt spatial integrity when processing grid-like data, our CSAC is explicitly designed with convolutional operations to preserve crucial spatial relationships throughout the network while simultaneously identifying and weighting significant temporal dependencies.

\noindent The main contributions of this work are:
\begin{itemize}[leftmargin=*]
    \item \textbf{A Novel BiDepth Mechanism:} We propose and implement a bidirectional depth modulation strategy within the network encoder. This allows for adaptive processing of ST sequences by varying analytical depth, enabling the model to effectively capture both long-term historical trends and short-term, immediate fluctuations.
    \item \textbf{A Spatially-Aware Convolutional Self-Attention Cell (CSAC):} We introduce a new CSAC that, distinct from many existing attention mechanisms, explicitly preserves spatial structures and relationships throughout the network, akin to standard convolutional operations, while effectively modeling temporal dependencies. This addresses a key limitation in applying attention to ST grid data.
    \item \textbf{The BiDepth Multimodal Neural Network (BDMNN):} We integrate these novel mechanisms into a cohesive, end-to-end deep learning architecture (BDMNN) that demonstrates superior ST forecasting performance on diverse real-world datasets when compared against leading benchmarks.
    \item \textbf{Efficient High-Dimensional Data Processing:} The BDMNN architecture incorporates a weight-sharing mechanism within its BiDepth component, which enhances parameter efficiency and model robustness, particularly when processing high-dimensional ST data.
\end{itemize}

The remainder of this paper is organized as follows. \Cref{sec:related_work} reviews related works in ST forecasting. \Cref{sec:methodology} details the proposed BDMNN architecture, including the BiDepth mechanism and the CSAC. \Cref{sec:experiments} presents the experimental setup, datasets, evaluation metrics, and comprehensive results with comparative analysis. Finally, \Cref{sec:conclusion} concludes the paper and outlines future research directions.

\section{Related Work}
\label{sec:related_work}

Numerous methods have been proposed for ST data prediction, broadly categorized into traditional statistical methods \citep{cressie2011statistics} and machine learning (ML) \citep{shi2018machine} or deep learning (DL) \citep{wang2020deep} approaches.

Traditional statistical models, such as the Auto-Regressive Integrated Moving Average (ARIMA) and its variants \citep{shekhar2007adaptive, li2012prediction, adhikari2013introductory, moreira2013predicting}, have been widely applied for ST prediction tasks. While some studies have incorporated spatial relations and external contextual data \citep{tong2017simpler, deng2016latent} to enhance time series predictions, these traditional models often struggle to capture complex ST correlations effectively, especially in high-dimensional, non-linear scenarios \citep{yao2018deep}. To specifically address ST correlations, multivariate Poisson log-normal models have been proposed \citep{xian2021spatiotemporal, yang2024measurement}. However, their reliance on predefined covariance structures and challenges with scalability can limit their applicability to complex, non-stationary ST data.

The advent of ML and DL has spurred significant progress in modeling complex ST correlations \citep{lecun2015deep}. Early ML methods applied to ST prediction include k-nearest neighbors (kNN) and support vector machines (SVMs). For example, \citet{arroyo2009forecasting} adapted kNN for forecasting histogram time series, and \citet{marchang2020knn} integrated ST correlations into kNN for missing data inference. Similarly, \citet{feng2018adaptive} employed a multi-kernel SVM with ST correlations for short-term traffic flow prediction. While these methods can capture certain ST dependencies, they are often sensitive to metric or kernel choices and generally face difficulties with non-stationary patterns and scalability as data dimensionality increases.

Deep learning methods have emerged as a more robust alternative for handling the complexities of ST data. Convolutional Neural Networks (CNNs) \citep{ali2022exploiting, casallas2022design}, including architectures like U-Net \citep{ronneberger2015u}, excel at capturing spatial patterns but typically lack inherent sensitivity to temporal dynamics \citep{he2019stcnn}. Conversely, Recurrent Neural Networks (RNNs) and Long Short-Term Memory networks (LSTMs) \citep{hochreiter1997long} effectively model temporal correlations \citep{zenkner2023flexible} but often face challenges in directly processing and integrating spatial information from grid-like data structures \citep{shi2015convolutional}.

To leverage the strengths of both, hybrid architectures have been proposed. CNN-RNN models, such as those by \citet{lu2020cnn}, integrate convolutional layers with recurrent units to capture both spatial and temporal dependencies. However, they may still struggle with very long-range temporal dependencies and intricate ST interactions. \citet{shi2015convolutional} introduced the ConvLSTM, which embeds convolutional operations within the LSTM cell structure, allowing it to learn ST features simultaneously. While an important step, ConvLSTM can still face challenges in efficiently modeling non-stationary patterns, particularly in balancing the influence of short-term fluctuations against long-term trends. More recent models like LSTNet \citep{lai2018modeling} and the Vision Transformer (ViViT) \citep{arnab2021vivit} employ CNN-LSTM hybrids or self-attention mechanisms. However, a common limitation in some of these approaches, such as the ViViT, is the flattening of spatial dimensions to fit sequence-based mechanisms, which can result in the loss of critical spatial correlations and undermine the ability to fully model intricate spatial dependencies. Other approaches \citep{salekin2021multimodal, zhang2022multi} combine different data types or models, but may not comprehensively address the complexity of non-stationary patterns in high-dimensional ST data while fully preserving spatial context.

\Cref{tb:methods_comparison} summarizes the capabilities of representative state-of-the-art methods in capturing key ST characteristics. The limitations highlighted motivate our development of BDMNN, which aims to overcome these challenges by explicitly focusing on adaptive temporal depth and spatially-aware attention.
\begin{table*}[ht]
    \caption{Comparison of ST prediction approaches with respect to key ST characteristics. Our proposed method (BDMNN) aims to address all listed characteristics.}
    \label{tb:methods_comparison}
    \centering
    \resizebox{\textwidth}{!}{%
      \begin{threeparttable}
        \begin{tabular}{lccccc}
          \toprule
          \multirow{2.5}{*}{\textbf{ST characteristic}} & \multicolumn{2}{c}{\textbf{Traditional Statistical Method}} & \multicolumn{2}{c}{\textbf{Traditional ML/DL-based Method}} & \multirow{2.5}{*}{\textbf{Proposed Method}} \\
          \cmidrule(lr){2-3} \cmidrule(lr){4-5}
          & \textbf{ARIMA-based} & \textbf{Poisson Process-based} & \textbf{ML-based} & \textbf{DL-based} & \\
          \midrule
          High dimensionality       & No & No & No            & \textbf{Yes}      & \textbf{Yes} \\
          Complex ST correlations   & No & \textbf{Yes} & \textbf{Yes}  & \textbf{Yes}      & \textbf{Yes} \\
          Non-stationary patterns   & No & No & No            & Partially\tnote{*} & \textbf{Yes} \\
          \bottomrule
        \end{tabular}
        \begin{tablenotes}
          \item[*] While many deep learning models aim to capture non-stationary patterns, they often struggle to effectively balance the influence of long-term trends against short-term fluctuations, a limitation that directly motivates our architecture.
        \end{tablenotes}
      \end{threeparttable}
    }
\end{table*}

\section{Methodology}
\label{sec:methodology}

\begin{figure}[ht!] 
    \centering 
    \includegraphics[width=80mm]{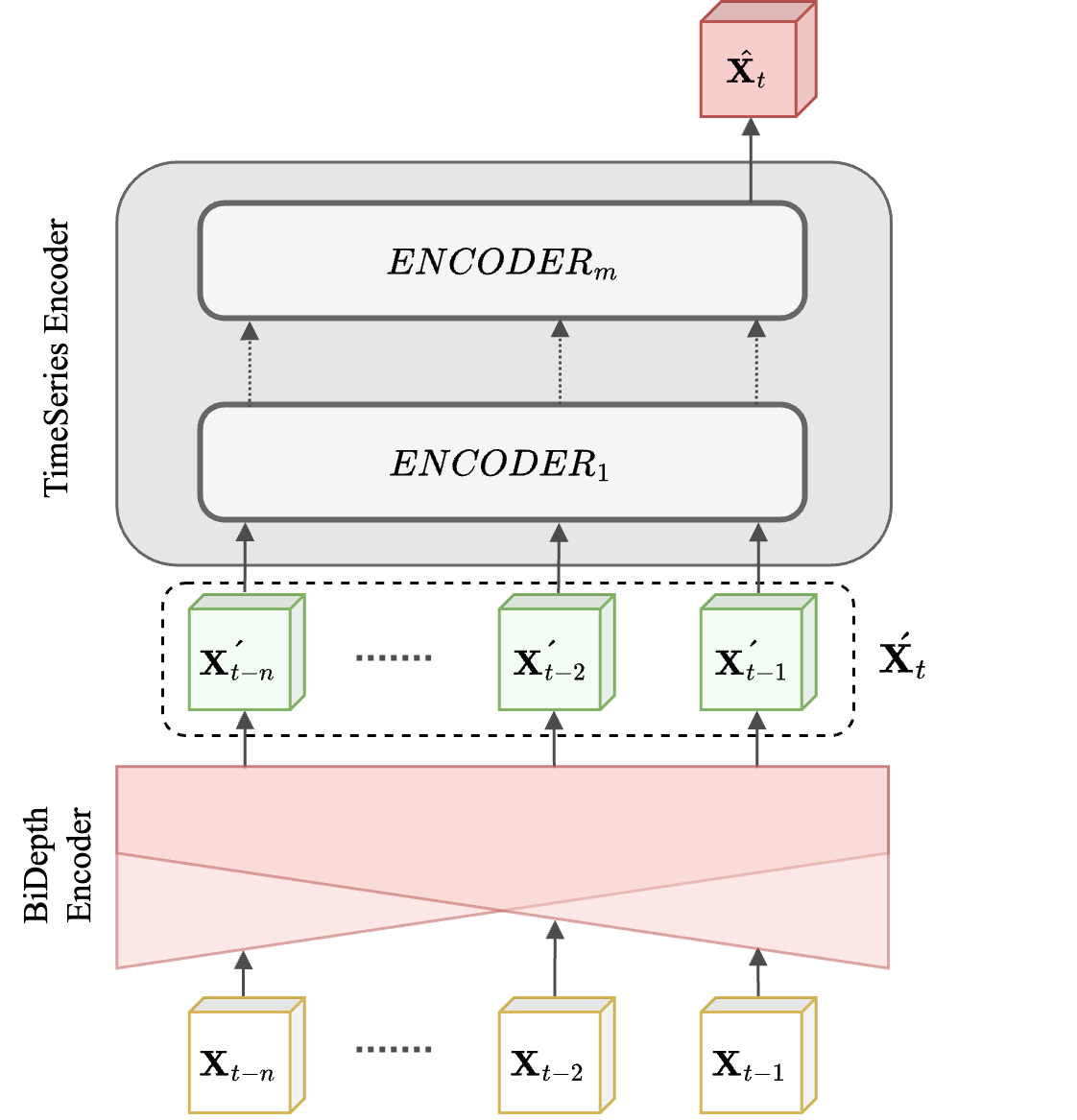} \caption{Overview of the BDMNN. The data is processed through the BiDepth Encoder, resulting in spatially transformed sequences. These sequences are then passed to the TimeSeries Encoder, which captures the temporal dynamics and produces the prediction for the target timestamp.} 
    \label{fig:BiDepthNet} 
\end{figure}

The BDMNN, delineated in this section, presents a novel method to navigate the intricate corridors of ST data analysis, addressing methodological gaps in the existing analytical models. Centralized around two pivotal components—the BiDepth Encoder and the TimeSeries Encoder — graphically illustrated in \Cref{fig:BiDepthNet}. Initially, we present a brief overview of the entire operation, followed by a detailed exploration of each individual component.

At the core of this framework is the BiDepth Encoder, which orchestrates the processing of each timestamp within a given sequence through convolutional layers. The Encoder integrates the BDMNN's dual aspects: a DeepShallow network and a ShallowDeep network. The DeepShallow network employs a decreasing layer depth along the temporal context, which reduces the layer depth for recent data and thus enables swift adaptation to new information, while the ShallowDeep network does the opposite, applying increased depth to recent data to capture complex, short-term dependencies. The rationale behind this BiDepth approach stems from the characteristics of ST data, where relationships between variables can vary in complexity over time and space. 
For instance, in the TLC data, taxi demand may have direct and straightforward relationships with recent past data at a local level—the demand in a specific neighborhood in the previous hour may closely predict the current demand. However, there may also be complex recent events, such as sudden traffic congestion due to an accident, that require a more complex model structure (i.e., deeper layers) to capture non-linear and intricate dependencies across a broader spatial area. Non-linear dependencies refer to relationships that cannot be captured by linear models, often arising from high-dimensionality and non-stationarity in the data. By using both decreasing and increasing layer depths, the model can capture simple, localized direct relations with shallow layers (simpler model structure) and more complex, overall spatial patterns with deeper layers.
Similarly, for longer-term dependencies, sometimes simple patterns like weekly seasonality (e.g., higher demand on Friday evenings in certain neighborhoods) can be captured with shallow layers focused on localized trends, while other times, more complex seasonal patterns (e.g., city-wide changes due to holidays or special events) require deeper layers to model effectively across the entire spatial domain. If we were to use only one directional depth assignment, we might miss capturing either the simple, localized relationships or the complex, global dependencies, leading to suboptimal performance. An equal depth approach may not provide the flexibility needed to adapt to varying complexities in the data over different time scales and spatial scales. Our experiments, as discussed in \Cref{sec:results}, demonstrate that the BiDepth approach outperforms single-directional (i.e. DeepShallow \citep{ehsani2024deepshallow} or ShallowDeep) or equal-depth architectures (i.e. ConvLSTM \citep{shi2015convolutional} or ViViT \citep{arnab2021vivit} ) in capturing the diverse temporal and spatial patterns present in ST data.

This combination of layer depth, alternating between the DeepShallow and ShallowDeep methodologies, allows the model to proficiently handle the intricate demands of ST data analysis. This flexible architecture not only adjusts to the varying complexity of the data but also ensures comprehensive processing of both historical and recent data, offering a nuanced understanding of the entire temporal spectrum.

As illustrated in \Cref{fig:BiDepthNet}, we represent our ST data as a tensor \(\mathbf{X} \in \mathbb{R}^{b \times n \times c \times h \times w}\). Here, \(b\) denotes the batch size, indicating the number of samples processed in parallel; \(n\) is the number of historical time steps (i.e., the window size), reflecting how many previous time intervals are considered, \(c\) represents the number of feature channels at each spatial location, such as 15-minute intervals capturing events like rush hours (e.g., \(c = 16\) for a 4-hour evening rush period). Finally, \(h, w\) correspond to the spatial height and width of the grid, where \(h\) and \(w\) partition the city into spatial regions (e.g., NYC taxi zones). This tensor format cohesively encapsulates ST data, including multiple time intervals, sequences of historical windows, and the spatial layout of the city.

Given a prediction time \( t \) and utilizing historical data spanning up to \( n \) time steps preceding \( t \) (represented as a window of size \( n \)), the historical inputs \(\mathbf{X}_{t-1}, \ldots, \mathbf{X}_{t-n}\) are processed by the BiDepth Encoder. Here, each \(\mathbf{X}_{t-i}\) represents the spatial data (e.g., taxi demands across all zones) at time \( t - i \). The BiDepth Encoder takes the entire historical window \(\{\mathbf{X}_{t-1}, \ldots, \mathbf{X}_{t-n}\}\) as input and learns the ST characteristics of the data, focusing on non-stationarity and complex spatial patterns. This results in a set of transformed sequences \(\{\mathbf{X}'_{t-1}, \ldots, \mathbf{X}'_{t-n}\}\), which encapsulate the learned spatial features and patterns. For convenience, we denote the collective transformed historical data as \(\mathbf{X}'_{t}\), where \(\mathbf{X}'_{t} = \{\mathbf{X}'_{t-1}, \ldots, \mathbf{X}'_{t-n}\}\). These transformed representations are then passed to the TimeSeries Encoder, which captures the temporal dependencies across the entire historical window to produce the final prediction. The comprehensive details of our BiDepth Encoder are presented in Section~\ref{subsec:bidepth_encoder}.

\begin{equation}
\mathbf{X}'_{t} = \text{BiDepthEncoder}(\mathbf{X}_{t-1}, \ldots, \mathbf{X}_{t-n}) .
\end{equation}

The TimeSeries Encoder offers flexibility in configuration, allowing for either a Convolutional LSTM or a Convolutional Self-Attention mechanism, each providing distinct advantages in modeling temporal dynamics. It predicts the values of \(\mathbf{X}_t\) based on the knowledge of historical ST patterns stored in \(\mathbf{X}'_{t}\). While the Convolutional LSTM accentuates long-term correlations (e.g., capturing seasonal trends), the Convolutional Self-Attention mechanism allows the model to dynamically identify and weight spatial correlations across the entire sequence. The culmination of this process is the prediction for time \( t \), denoted as \(\hat{\mathbf{X}}_{t}\). The details of the TimeSeries Encoder are introduced in Section~\ref{sec:temporal_encoder}.

\begin{equation}
\hat{\mathbf{X}}_{t} = \text{TimeSeriesEncoder}(\mathbf{X}'_{t}).
\end{equation}

In synergy, the BiDepth Encoder and TimeSeries Encoder craft a robust model for ST data processing, harmoniously merging computational efficiency with nuanced model complexity. The following sections undertake a detailed dissection of each component's blueprint and functionality.

\subsection{BiDepth Encoder}
\label{subsec:bidepth_encoder}

\begin{figure}[ht!]
  \centering
  \includegraphics[width=\textwidth]{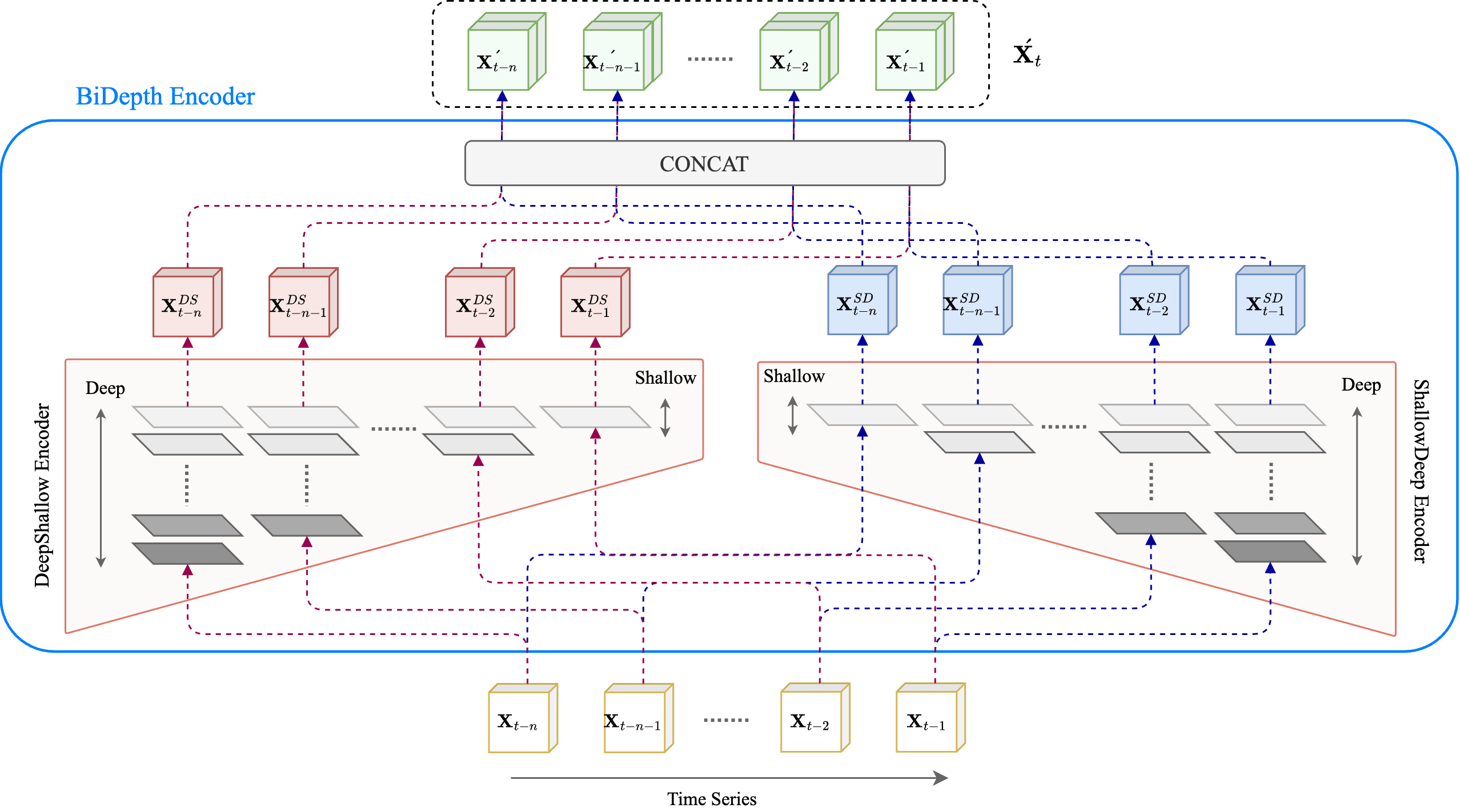}
  \caption{Schematic Representation of the BiDepth Encoder Architecture. The BiDepth Encoder consists of two main components: the DeepShallow Encoder and the ShallowDeep Encoder. These encoders analyze time series data differently; the DeepShallow Encoder focuses on deeper layers for earlier time steps and shallower layers for more recent time steps, while the ShallowDeep Encoder does the opposite, emphasizing deeper layers for recent data and shallower layers for older data. Each time component of the series is processed by both encoders, with the results concatenated at the output. The architecture employs a weight-sharing strategy where Convolutional Neural Networks illustrated as plates) at corresponding depths across the encoders share weights, indicated by the identical shading. This approach facilitates the effective capture of both ST dependencies, enhancing the model's capability to handle complex ST data.}
    \label{fig:DeepShallow}
\end{figure}

The BiDepth Encoder, illustrated in \Cref{fig:DeepShallow}, is designed to integrate the capabilities of the two complementary networks: the DeepShallow and ShallowDeep encoders. This encoder serves as the backbone for processing ST data, ensuring a comprehensive and balanced analysis.

Each input, \(\mathbf{X}_{t-i}\) at a specific historical timestamp, \(t-i\), is processed through both DeepShallow and ShallowDeep networks. In the context of our datasets, this input represents the spatial data at time \( t - i \), such as the taxi demand across all zones in the TLC dataset or the precipitation measurements across all grid points in the GPM dataset (described in \Cref{sec:gpm_data}). After processing, the BiDepth Encoder combines the outputs of these two encoders. This concatenation step \citep{szegedy2015going} is pivotal, as it merges insights into a unified representation. The concatenated output is then fed into the TimeSeries Encoder of the BDMNN for subsequent processing and forecasting.

\begin{equation}
\mathbf{X}'_{t-i} = \text{Concat}(\mathbf{X}_{t-i}^{DS}, \mathbf{X}_{t-i}^{SD}), \quad \text{for } i = 1, \ldots, n \hspace{0.5em} ,
\end{equation}

\noindent where \( \mathbf{X}'_{t-i} \) represents the output of the BiDepth Encoder at timestamp \( t - i \), and \( \mathbf{X}_{t-i}^{DS} \) and \( \mathbf{X}_{t-i}^{SD} \) are the outputs of the DeepShallow and ShallowDeep encoders, respectively, at the same timestamp \( t - i \). The construction of the network can be summarized in \Cref{alg:BiDepth}.

\begin{algorithm}[ht!]
\caption{BiDepth Network Algorithm}
\begin{algorithmic}[1]
\Require Input tensor $\mathbf{X} \in \mathbb{R}^{b \times n \times c_{\text{in}} \times h \times w}$, initial maximum depth $L$, window size $n$
\Ensure Output tensor $\mathbf{\hat{X}}_{t} \in \mathbb{R}^{b \times c_{\text{out}} \times h \times w}$

\State \textbf{Initialize:}
\State \quad Convolutional layers $\{\text{CNN}_d\}_{d=1}^{L}$ shared across depths
\State \quad Compute layer decrement $\delta = \dfrac{L - 1}{n - 1}$

\For{$i = 1$ to $n$} 
    \Statex \hspace{1.2em} \textit{\# DeepShallow (DS) Encoder:}
    \State Compute depth for DS: $D^{DS}_{(t-i)} = \max\left(1, L - \left\lfloor (i - 1) \delta \right\rceil \right)$
    \For{$d = 1$ to $D^{DS}_{(t-i)}$}
        \State $\mathbf{X}^{DS}_{t-i} \leftarrow \text{ReLU}\left(\text{CNN}_d\left(\mathbf{X}^{DS}_{t-i}\right)\right)$
    \EndFor

    \Statex \hspace{1.2em}\textit{\# ShallowDeep (SD) Encoder:}
    \State Compute depth for SD: $D^{SD}_{(t-i)} = \max\left(1, \left\lfloor (i - 1) \delta \right\rceil + 1\right)$
    \For{$d = 1$ to $D^{SD}_{(t-i)}$}
        \State $\mathbf{X}^{SD}_{t-i} \leftarrow \text{ReLU}\left(\text{CNN}_d\left(\mathbf{X}^{SD}_{t-i}\right)\right)$
    \EndFor

\EndFor

\Statex \textit{\# Stack outputs over time:}
\State \quad $\mathbf{X}^{DS}_t \leftarrow \text{Stack}\left(\mathbf{X}^{DS}_{t-1}, \mathbf{X}^{DS}_{t-2}, \dots, \mathbf{X}^{DS}_{t-n}\right)$
\State \quad $\mathbf{X}^{SD}_t \leftarrow \text{Stack}\left(\mathbf{X}^{SD}_{t-1}, \mathbf{X}^{SD}_{t-2}, \dots, \mathbf{X}^{SD}_{t-n}\right)$

\Statex \textit{\# Concatenate paths along channel dimension:}
\State \quad $\mathbf{X}'_t \leftarrow \text{Concat}\left(\mathbf{X}^{DS}_t, \mathbf{X}^{SD}_t\right)$

\Statex \textit{\# Apply ConvLSTM or ConvSelfAttention:}
\State \quad $\mathbf{\hat{X}}_t \leftarrow \text{TimeSeriesEncoder}\left(\mathbf{X}'_t\right)$

\State \textbf{Return} $\mathbf{\hat{X}}_t$

\end{algorithmic}
\label{alg:BiDepth}
\end{algorithm}

\subsubsection{DeepShallow Encoder}

The DeepShallow encoder is characterized by two key architectural features: (i) a dynamic adjustment of CNN layer depths as time progresses and (ii) a shared-weight mechanism for layers exhibiting identical depth. We focus on these two features because they are essential for adapting the model to temporal dynamics and improving computational efficiency. These features distinguish our approach from traditional methods that often use fixed-depth layers without weight sharing.

At its core, the DeepShallow encoder processes sequences by channeling each timestamp through a set of CNN layers, as illustrated in \Cref{fig:DeepShallow}. Unlike conventional models, these layers are dynamic in depth, changing based on the timestamp's temporal position. The rationale behind varying the layer depth is to capture the varying complexity of patterns over different time scales. For earlier (older) timestamps, deeper layers can capture complex long-term dependencies, while shallower layers (fewer layers) are sufficient for more recent timestamps where patterns may be simpler or more directly related to the prediction target.

This feature is adjusted by a user-defined function, \textit{depth\_function}, potentially a linear function, that orchestrates the depth transition throughout the sequence. The depth function is mathematically represented as shown in Equation~(\ref{eq:depthFunction}), where \(D^{DS}_{(t-i)}\) indicates the number of convolutional layers (i.e., the depth) applied at \( i \) timestamps prior to \( t \):

\begin{equation}
D^{DS}_{(t-i)} = \max\left(1, L - \left\lfloor \frac{(i - 1)(L - 1)}{n - 1} \right\rceil \right), \quad \text{for } i = 1, \ldots, n \hspace{0.5em}.
\label{eq:depthFunction}
\end{equation}

\noindent In this function, \( L \) is the initial maximum depth, \( n \) is the window size (number of historical timestamps), and \( i \) indexes the timestamps from the most recent (\( i = 1 \)) to the oldest (\( i = n \)). The notation \(\left\lfloor \right\rceil\) indicates rounding to the nearest integer. The depth decreases linearly from \( L \) for the oldest timestamp to 1 for the most recent. Alternative depth functions can be used, but we found the linear function effective in balancing complexity and efficiency. We selected this linear function as it provides a simple, interpretable, and parameter-free baseline to robustly demonstrate the core principle of the BiDepth mechanism. Investigating more complex, learnable depth functions is left as a promising direction for future work.

The number of convolutional layers \citep{lecun1998gradient} applied to each timestamp \( t-i \) is given by \(D^{DS}_{(t-i)}\). Each depth level \( d = 1, \ldots, D^{DS}_{(t-i)} \) corresponds to one convolutional layer. To ensure efficiency and prevent overfitting, layers at the same depth level share their weights across all timestamps. This means that if two timestamps use a layer of depth \( d \), they apply the same convolutional filters \( W_d \). As \(D^{DS}_{(t-i)}\) varies over time, certain timestamps may involve fewer layers (shallow configurations) and others more layers (deeper configurations), but layers at the same depth index \( d \) always use the same set of shared weights \( W_d \).

Formally, given the shared weights \( W_{d} \) for each depth level \( d \), the output \(\mathbf{X}^{DS}_{t-i}\) for timestamp \( t-i \) is:

\begin{equation}
\mathbf{X}^{DS}_{t-i}  = \text{CNN}_{D^{DS}_{(t-i)}}\left( \mathbf{X}_{t-i}; \{W_{d}\}_{d=1}^{D^{DS}_{(t-i)}} \right), \quad \text{for } i = 1, \ldots, n \hspace{0.5em},
\label{eq:cnnOperationDS}
\end{equation}

\noindent where \(\text{CNN}_{D^{DS}_{(t-i)}}(\cdot)\) indicates applying \(D^{DS}_{(t-i)}\) convolutional layers in sequence, each layer identified by a distinct depth index \( d \). The set \(\{W_{d}\}_{d=1}^{D^{DS}_{(t-i)}}\) denotes the weights for all layers up to the current depth. Layers with the same depth index \( d \) use the same weights \( W_{d} \) at every timestamp where that depth is employed, ensuring consistent feature extraction across different temporal contexts. By sharing weights among layers of the same depth, the model reduces the total number of parameters and learns generalized features that remain effective across varying time steps.

\subsubsection{ShallowDeep Encoder}

The ShallowDeep encoder adopts an inverse strategy compared to the DeepShallow encoder. Instead of decreasing layer depth over time, it increases the number of convolutional layers for more recent timestamps, focusing on short-term, complex patterns. This approach allows for more detailed modeling of recent data, where abrupt changes and localized events may require a higher number of convolutional layers. The depth function for the ShallowDeep encoder is defined as:

\begin{equation}
D^{SD}_{(t-i)} = \max\left(1, \left\lfloor \frac{(i - 1)(L - 1)}{n - 1} \right\rceil + 1 \right), \quad \text{for } i = 1, \ldots, n \hspace{0.5em},
\label{eq:depthFunctionSD}
\end{equation}

\noindent where \(D^{SD}_{(t-i)}\) specifies how many convolutional layers are applied at timestamp \( t - i \). The depth increases from 1 for the oldest timestamp to \( L \) for the most recent, reflecting the need for deeper modeling of current short-term dependencies. Similar to the DeepShallow encoder, the ShallowDeep encoder employs a weight-sharing mechanism where all layers at the same depth index share a common set of weights \( W_{d} \). Given the shared weights for each depth level, the output \(\mathbf{X}^{SD}_{t-i}\) is:

\begin{equation}
\mathbf{X}^{SD}_{t-i} = \text{CNN}_{D^{SD}_{(t-i)}}\left( \mathbf{X}_{t-i}; \{W_{d}\}_{d=1}^{D^{SD}_{(t-i)}} \right), \quad \text{for } i = 1, \ldots, n \hspace{0.5em}.
\label{eq:cnnOperationSD}
\end{equation}

\noindent In this operation, \(\text{CNN}_{D^{SD}_{(t-i)}}(\cdot)\) applies \(D^{SD}_{(t-i)}\) convolutional layers in sequence, each identified by a depth index \( d \). Layers sharing the same depth index use the same weights \( W_{d} \) across all timestamps, ensuring consistent feature extraction and efficient parameter utilization.

\subsection{TimeSeries Encoder}
\label{sec:temporal_encoder}

The TimeSeries Encoder is responsible for modeling the temporal dependencies in the transformed spatial representations \(\mathbf{X}'_t\) produced by the BiDepth Encoder. It takes \(\mathbf{X}'_t \in \mathbb{R}^{b \times n \times c \times h \times w}\) as input, where the dimensions \(b, n, c, h, w\) follow the definitions provided in the BiDepth Encoder section.

While our framework can employ various temporal modeling approaches, such as a Convolutional LSTM \citep{shi2015convolutional}, our primary contribution lies in the development of a Convolutional Transformer-based TimeSeries Encoder. This design leverages CASC to capture temporal dependencies without discarding spatial information. The CASC developed in our work addresses the challenge of effectively retaining spatial information during the application of attention mechanisms \citep{guo2019attention}. Our experimental results (see Section~\ref{sec:results}) indicate that this Convolutional Transformer approach outperforms the ConvLSTM baseline, particularly in handling complex ST correlations and multi-scale dynamics.

\begin{figure}[ht!]
  \centering
  \includegraphics[width=\textwidth]{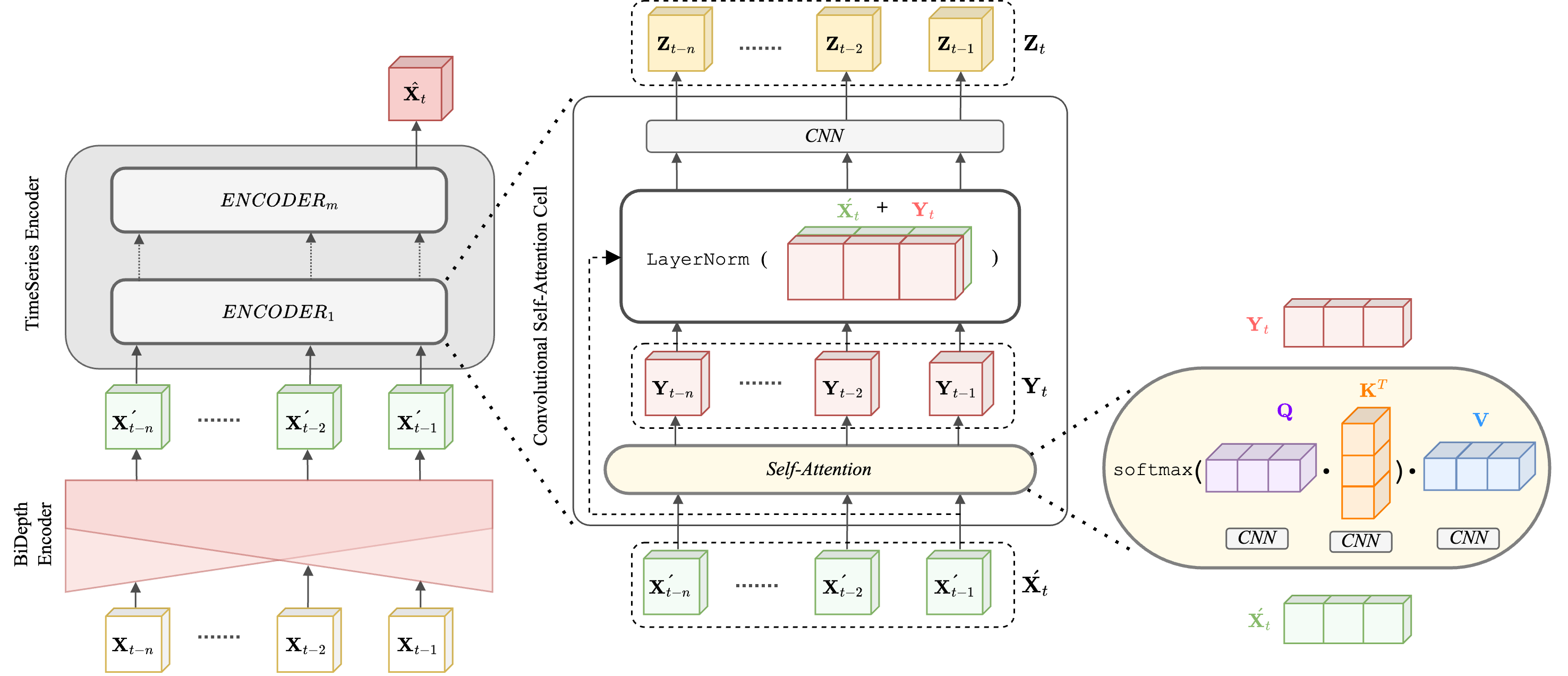}
  \caption{Illustration of the Convolutional Self-Attention Cell. The cell's internal workflow is displayed, emphasizing the processing of the input tensor through the Self-Attention layer. This layer employs 2D convolutional layers to generate the query, key, and value tensors for calculating attention weights. Subsequent to this, the attention mechanism modifies the value tensor, yielding an output that is conscious of other timestamps. The final output merges the input tensor with these attention-informed tensors, passing through layer normalization and an additional 2D convolution layer. This process exemplifies our model's ability to seamlessly incorporate the attention mechanism while safeguarding the input data's spatial structure.}
  \label{fig:AttentionCell}
\end{figure}

The CSAC, as illustrated in Figure~\ref{fig:AttentionCell}, extends the self-attention mechanism by preserving spatial structures through convolutional operations. Let \(\mathbf{X}'_t \in \mathbb{R}^{b \times n \times c \times h \times w}\) be the input sequence of transformed spatial features over \(n\) time steps. We employ three separate convolutional operations to generate query (\(\mathbf{Q}\)), key (\(\mathbf{K}\)), and value (\(\mathbf{V}\)) tensors from \(\mathbf{X}'_t\). Formally, we define:

\begin{equation}
\mathbf{Q} = \text{CNN}(\mathbf{X}'_t; W_Q), \quad
\mathbf{K} = \text{CNN}(\mathbf{X}'_t; W_K), \quad
\mathbf{V} = \text{CNN}(\mathbf{X}'_t; W_V) ,
\end{equation}

\noindent where $\mathbf{Q}$, $\mathbf{K}$, and $\mathbf{V}$ represent convolutional mappings \citep{lecun1998gradient}, each parameterized by its own weights (\(W_Q, W_K, W_V\)). These convolutions retain the spatial dimensions \((h,w)\), ensuring that spatial relationships are not lost. This approach is fundamentally different from standard attention mechanisms that typically flatten the $h$ and $w$ spatial dimensions into a single sequence, an operation that disrupts the inherent grid structure and loses critical spatial relationships.

After reshaping \(\mathbf{Q}\) and \(\mathbf{K}\) for the attention computation, the attention scores \(\mathbf{A}\) are obtained via a softmax operation applied to \(\mathbf{Q}\mathbf{K}^\text{T}\) (where \(\mathbf{K}^\text{T}\) denotes the transpose of \(\mathbf{K}\)):

\begin{equation}
\mathbf{A} = \text{softmax}(\mathbf{Q}\mathbf{K}^\text{T}) ,
\end{equation}

\noindent  \(\mathbf{A} \in \mathbb{R}^{b \times n \times n}\) represents the temporal attention weights, indicating how each time step in the sequence attends to every other time step. Applying these attention weights to \(\mathbf{V}\) yields the attention-informed output \(\mathbf{Y}_t\):

\begin{equation}
\mathbf{Y}_t = \mathbf{A}\mathbf{V} ,
\end{equation}

\noindent 
\(\mathbf{Y}_t\) shares the same shape as \(\mathbf{X}'_t\), i.e., \(\mathbf{Y}_t \in \mathbb{R}^{b \times n \times c \times h \times w}\), ensuring that the output remains spatially coherent. Here, \(\mathbf{A}\) encodes temporal correlations between different time steps, and by applying \(\mathbf{A}\) to \(\mathbf{V}\), we produce \(\mathbf{Y}_t\) that emphasizes relevant temporal patterns across the entire sequence.

In the final integration step, we concatenate \citep{szegedy2015going} the original input \(\mathbf{X}'_t\) with the attention-informed output \(\mathbf{Y}_t\), forming \(\text{concat}(\mathbf{X}'_t, \mathbf{Y}_t)\). This concatenated tensor is then normalized using Layer Normalization and passed through an additional 2D convolutional layer:

{
\begin{align}
\mathbf{Y}_t^{\text{norm}} &= \text{LayerNorm}(\text{Concat}(\mathbf{X}'_t, \mathbf{Y}_t)), \\
\mathbf{Z}_t &= \text{CNN}(\mathbf{Y}_t^{\text{norm}}; W_O),
\end{align}
}

\noindent  where \(\mathbf{Z}_t \in \mathbb{R}^{b \times n \times c \times h \times w}\) represents the final output of the TimeSeries Encoder, combining the original input features with the attention-derived insights. \(\text{CNN}(\cdot; W_O)\) is another convolutional operation \citep{lecun1998gradient} with its own learnable weights \(W_O\), ensuring a flexible representation that can adapt to various ST complexities.

The Convolutional Transformer-based TimeSeries Encoder integrates convolutional operations within the self-attention mechanism to preserve spatial integrity, adaptively focus on temporal dependencies, and produce a refined output \(\mathbf{Z}_t\). This approach provides a more expressive modeling framework than the ConvLSTM, as evidenced by our empirical evaluations, and can be extended to multiheaded or multilayered structures for capturing increasingly complex spatial and temporal relationships.

\subsection{Evaluation Metrics}
\label{sec:Eval}

Our primary quantitative measures for model assessment were the Mean Squared Error (MSE) and the Mean Absolute Error (MAE). Both metrics are widely used in regression tasks, offering complementary insights into the model’s predictive performance \citep{pan2024reliability}. The MSE is defined as:

\begin{equation}
MSE = \frac{1}{s} \sum_{i=1}^{s} \frac{1}{h \times w} \sum_{j=1}^{h} \sum_{k=1}^{w} (X_{i,j,k} - \hat{X}_{i,j,k})^2 ,
\end{equation}

\noindent where \({X}_{i,j,k}\) and \(\hat{{X}}_{i,j,k}\) denote the true and predicted values at spatial coordinates \((j,k)\) for the \(i\)-th sample, and \(s, h, w\) represent the number of samples, and the spatial dimensions of the prediction respectively. MSE evaluates the average squared error and heavily penalizes larger deviations. This makes it valuable for identifying when the model significantly under- or overestimates the target, thus providing a sensitive gauge of model stability and performance.

MAE measures the average absolute error in the same units as the predicted quantity, offering a direct and intuitive interpretation of the model’s typical error magnitude. This linear treatment of errors makes MAE particularly useful for interpreting spatial distributions of prediction errors. For instance, an MAE of 4 for a zone in TLC data can be directly understood as being off by about 4 passengers on average. Such interpretability aids in practical decision-making and in understanding where and by how much the model tends to mispredict. The MAE is defined as:

\begin{equation}
MAE = \frac{1}{s} \sum_{i=1}^{s} \frac{1}{h \times w} \sum_{j=1}^{h} \sum_{k=1}^{w} |{X}_{i,j,k} - \hat{{X}}_{i,j,k}| .
\end{equation}

\subsection{Model Hyperparameter Tuning}

The performance of the BDMNN depends on several hyperparameters that control the architecture and training process. Key hyperparameters include the initial depth of the convolutional layers, the window size (number of historical timestamps used), the choice of activation functions, learning rate, batch size, and the number of filters in the convolutional layers.
Among these, the initial depth of the convolutional layers ($L$) in the BiDepth Encoder is a significant hyperparameter that directly affects the model's capacity to capture both historical and recent trends in the data. The initial depth ($L$) determines the maximum number of convolutional layers through which the data passes at each timestamp, as governed by the depth functions defined in Equations~(\ref{eq:depthFunction}) and (\ref{eq:depthFunctionSD}).

Our analysis has shown that increasing $L$ can lead to improved model performance up to a certain point. Specifically, as the depth increases, the MSE on the validation set decreases, indicating better predictive accuracy. However, this comes at the cost of increased computational complexity and longer training times. There is a trade-off between model depth, predictive performance, and computational cost. An optimal depth exists where the MSE is minimized without incurring unnecessary computational overhead. Selecting this optimal depth requires balancing the benefits of increased model capacity with the practical considerations of training time and resource availability.

In our experiments, detailed in Section~\ref{sec:ImpactOfDepth}, we empirically evaluate the impact of the initial depth on the BDMNN's performance using the TLC data. The results corroborate the conceptual understanding, showing that beyond a certain depth, further increases do not significantly improve MSE but do increase training time.




\section{Case Studies}
\label{sec:experiments}

To demonstrate the effectiveness of the proposed BDMNN, we conducted two case studies with two different ST datasets:  the NYC TLC data and the GPM datasets. By leveraging these datasets, we aim to validate our model's versatility across diverse complexities and applicative domains. The NYC TLC data preprocessing details can be found in Subsection~\ref{sec:data_processing}. \Cref{tab:datasets} summarizes the key ST characteristics of the two datasets, highlighting high dimensionality, ST correlation, and evolving non-stationary patterns.



\begin{table*}[ht]
\caption{ST Characteristics of the Datasets}
\label{tab:datasets} 
\centering
\resizebox{1\columnwidth}{!}{
\begin{tabular}
{>{\centering\arraybackslash}m{3cm}|>{\raggedright\arraybackslash}m{7cm}|>{\raggedright\arraybackslash}m{7cm}}
\hline\hline
\textbf{ST Characteristic} & \textbf{TLC Dataset} & \textbf{GPM Dataset}  \\ \hline
\textbf{High Dimensionality} & 
\begin{itemize}
    \item Over 200 taxi zones
    \item 15-min intervals 
    \item Over three years
\end{itemize}
 & 
 \begin{itemize}
     \item Spatial resolution of 0.1°
     \item Covering continental US
     \item 30-min precipitation data 
     \item Over 6 years
 \end{itemize} 
\\\hline
\textbf{Complex ST Correlations} &
\begin{itemize}
    \item Taxi demand in one zone impacted by adjacent zones and recent trends
\end{itemize}
  & 
  \begin{itemize}
      \item Precipitation at a grid affected by neighboring grids and temporal evolution of weather systems
  \end{itemize}
\\ \hline
\textbf{Non-Stationary Patterns} &  
\begin{itemize}
    \item Short-term spikes during rush hours or disruptions
    \item Long-term seasonal trends like higher demand during holidays or summer
\end{itemize}& 
\begin{itemize}
    \item Short-term events like sudden storms or flash floods
    \item Long-term seasonal trends such as monsoon rainfall increases
\end{itemize}
 \\ \hline\hline
\end{tabular}}
\end{table*}

\subsection{Data Preprocessing}
\label{sec:data_processing}

\subsubsection{TLC Data}
\label{sec:tlc_data}
Originating from the NYC TLC, the TLC Trip Record Data offers an exhaustive snapshot of taxi trips in New York City \citep{taxi2017tlc}. The data, sourced from authorized technology providers under the Taxicab \& Livery Passenger Enhancement Programs (TPEP/LPEP), encompasses attributes such as pick-up and drop-off dates/times, locations, trip distances, itemized fares, rate types, payment types, and driver-reported passenger counts. Despite TLC's continuous reviews to ensure completeness and accuracy, the published data may not encapsulate all trips dispatched by TLC-licensed bases, highlighting the vast and complex nature of urban transport data. This dataset spans from 2009 to 2023, offering a comprehensive temporal landscape to evaluate the BDMNN. However, leveraging this dataset for our BDMNN required transforming it into a format that captures both ST dynamics. Consequently, our primary aim was to sculpt a ST tensor representing the taxi traffic patterns for a day based on the data from the preceding day.

\begin{figure}[ht!]
  \centering
  \includegraphics[width=\textwidth]{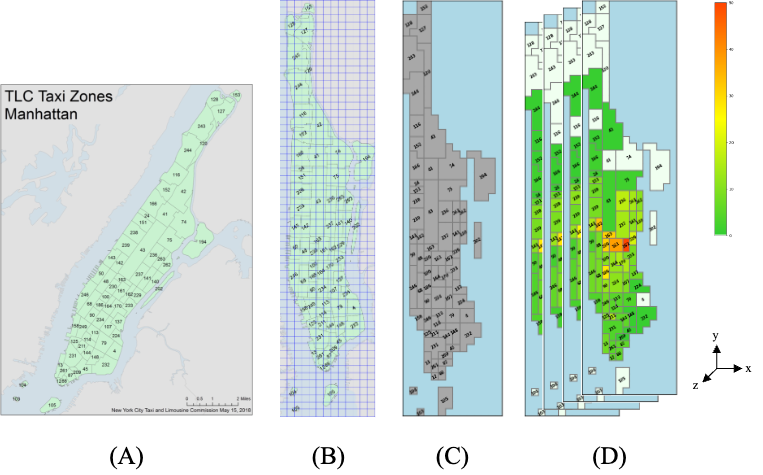}
  \caption{Sequential depiction of the New York City taxi zone processing. (A) Original TLC map marking various taxi zones in NYC. (B) After rotation for optimal tensor representation, the map showcases a grid layout matched to the smallest taxi zone's dimensions. This assures a uniform and detailed citywide spatial segmentation. (C) The post-pixelation view where each grid cell's value is ascertained using a max pooling approach, correlating it with the dominant zone. (D) The 3D tensor visualization conveys the transformation of pickup data into a ST structure, where x and y dimensions are spatial coordinates across NYC, and the z-dimension signifies time. Each tensor cell value indicates the passenger pickups in a grid segment during a specific 15-minute window, converting the city map into a succinct ST representation.}
    \label{fig:NYCGrid}
\end{figure}

The dataset was divided into a training set that includes data from 2021 and 2022, and a testing and validation set that contains data from 2023. By allocating older data for training and more recent data for validation and testing, we simulate a realistic forecasting scenario where the model predicts future patterns without access to future information during training. This chronological division reduces the risk of the model overfitting to temporal trends observed in future data, thereby ensuring a more unbiased and reliable model assessment. For a rigorous model evaluation, we split the 2023 data into two: the initial half (January to March) served as the validation set for model parameter tuning and optimization, while the latter half (April to May) functioned as the testing set. 

Concentrating on essential features of the dataset, such as pickup and drop-off times, locations, and passenger count, we focused on two critical periods to capture the pulse of urban transportation: the evening rush hour from 4 PM to 6 PM and the morning hours from 6 AM to 10 AM. The pickup data during the evening rush hour was aggregated into 15-minute intervals, crafting a temporal sequence of passenger pickups. Concurrently, we transformed the morning drop-off data similarly, providing another perspective on taxi service distribution and density.

Addressing the spatial dimension was challenging due to the dataset's discrete location identifiers. Our remedy was the provided NYC taxi zone map (Figure~\ref{fig:NYCGrid}A-C). The NYC taxi dataset divides the city into approximately 200 zones. We identified the smallest zone (i.e., the one with the least area) and used its dimensions to define a uniform grid overlaying the entire city. Each grid cell’s zone identity is determined by the zone covering the largest fraction of that cell's area, effectively “assigning precedence” to the dominant zone within that cell. This approach, inspired by the 'max pooling' concept in image processing, creates a grid-based representation of zones. While graph-based representations offer an alternative for irregular zones, we selected a grid-based approach for its computational efficiency and seamless compatibility with the convolutional operations central to our architecture.

We validated the spatial representation accuracy by comparing the actual area proportions of Manhattan zones to their "pixelated" (grid-based) counterparts. The average error rate between these two representations, weighted by the demand (average passenger count per location for each 15-minute interval), was found to be approximately 0.1. To calculate this, we first determined the area proportion of each zone in Manhattan based on the actual area and the grids, then computed the absolute difference between these two proportions for each zone. These differences were weighted by the respective location demand before averaging. For context, the mean demand per zone per 15-minute interval in our dataset is around 50 passengers, with a standard deviation of about 15 passengers. An average misallocation of 0.1 passengers implies that the spatial discretization introduced by the grid mapping minimally affects the overall demand distribution. The range of weighted errors varied, with a minimum of 0.00019, a maximum of 1.11689, and a standard deviation of 0.17495, reflecting the distribution of demand across zones. Since this error pertains to spatial relationships and typically occurs between neighboring zones, the impact of such misallocation on modeling performance is minimal. 

Following this spatial structuring, we integrated the temporally formatted pickup and drop-off data, converting our 2D grid into a 3D tensor. This tensor combined two spatial dimensions and a third temporal dimension, marked by successive 15-minute intervals. Within this tensor, each cell value conveys the passenger pickups for a specific zone during a designated time slot, crafting a dynamic traffic heatmap of New York City (Figure~\ref{fig:NYCGrid}D). In this heatmap, colors represent the intensity of demand, with warmer colors typically indicating higher passenger counts and cooler colors indicating lower counts. We then focused on Manhattan, presenting a time-sensitive, spatially granular depiction of passenger demands in this area.

\subsubsection{GPM Data}
\label{sec:gpm_data}

We utilized the GPM dataset described by \citet{ehsani2022nowcasting} in their NowCasting research. This dataset spans six years (2015-2020) and consists of the Global Precipitation Measurement (GPM) Integrated Multi-satellite Retrievals for GPM (IMERG) early run products \citep{huffman2020integrated}. It offers half-hourly precipitation maps at a spatial resolution of 0.1° across the Continental United States (CONUS).

The dataset was divided into training (85\%), validation (5\%), and testing (10\%) sets based on chronological order, with the most recent 10\% reserved for testing. This approach ensures the model is trained on past precipitation patterns and evaluated on future periods, mirroring real-world forecasting conditions. Throughout the training period, the model consistently uses a 24-hour historical window to predict the subsequent 4 hours of precipitation, maintaining a uniform training strategy over the entire training dataset.


\subsection{Computational Setup}
All experiments were conducted on NVIDIA A100 GPUs to ensure consistency across different datasets. The BDMNN and all baseline models were implemented using PyTorch. We employed hyperparameter tuning for all experiments to optimize model performance. Key hyperparameters included a learning rate of 0.005 for the AdamW optimizer and the use of MSE as the loss function. We also implemented early stopping with patience to prevent overfitting and improve training efficiency. The training time varied depending on the dataset, with an average of approximately 320 minutes for the GPM data and 8 minutes for the TLC data. To ensure reproducibility and assess model stability, we ran each model configuration with different random seeds at least 5 times, reporting the mean performance.


\subsection{Results}
\label{sec:results}

The evaluation utilized the MSE loss mentioned in \Cref{sec:Eval}. This section aims to delve into the performance of the BDMNN and its variants, providing insights into the model's capabilities across the datasets. To assess the individual contributions of the core components of our proposed BDMNN architecture, particularly the DeepShallow, ShallowDeep, and combined BiDepthNet encoders, we present an ablation study. These components were added to two distinct TimeSeries Encoder backbones: a standard ConvLSTM and our proposed ConvSelfAttention mechanism. The results are detailed below.

\subsubsection{TLC Trip Record Prediction}

For the TLC data, \Cref{tab:results_tlc_bidepth} showcases the performance of different models, emphasizing the improvements realized by incorporating the BiDepthNet structure.

\begin{table}[ht!]
\centering
\begin{tabular}{lccc}
    \toprule
    \textbf{Model} & \textbf{Model Complexity}\tnote{$\ast$}  & \textbf{MSE ($\pm$ Std. Dev.)} & \textbf{Improvements} (\%) \\
    \midrule
    ARIMA \citep{dubey2021study} & -   & $6.455 \pm 0.000$ & - \\
    SARIMA  \citep{dubey2021study} & - & $5.517 \pm 0.000$ & - \\
    U-Net \citep{ronneberger2015u} & 2,000 & $3.021 \pm 0.051$ & -   \\
    CNN-RNN  \citep{lu2020cnn} & 2,400 & $2.305 \pm 0.143$ & -  \\
    ViViT \citep{arnab2021vivit} & 460 & $1.831 \pm 0.021$ & - \\
    \hdashline
    ConvLSTM  \citep{shi2015convolutional}   & 560   & $1.652 \pm 0.012$ & -   \\
    + DeepShallow & 600 & $1.515 \pm 0.025$ & +9.87 \\
    + ShallowDeep & 600 & $1.500 \pm 0.022$ & +9.19   \\
    + BiDepthNet & 660 & $1.479 \pm 0.032$ & +10.47 \\
    \hdashline
    ConvSelfAttention & 350 & $1.621 \pm 0.041$ & - \\
    + DeepShallow & 380   & $1.503 \pm 0.038$ & +7.27 \\
    + ShallowDeep & 380 & $1.580 \pm 0.044$ &  +2.51 \\
    + BiDepthNet & 440 & $\mathbf{1.428 \pm 0.039}$ &  +11.89 \\
  \bottomrule
\end{tabular}
\caption[Performance Comparison of BiDepth vs Benchmarks on TLC Data]{
Performance comparison and model complexity of BiDepth versus Benchmarks on TLC Data. The performance depicted is based on MSE loss (mean $\pm$ standard deviation from 5 runs). The model parameters are presented in thousands. Models with modifications, such as “+ DeepShallow,” “+ ShallowDeep,” and “+ BiDepthNet,” denote enhancements made to the foundational models (TimeSeries Encoders: ConvLSTM and ConvSelfAttention). The reported improvements correspond to the percentage increase in performance (decrease in MSE) achieved by incorporating the introduced BiDepth encoder components. \\
{ \footnotesize $\ast$ Model complexity is assessed based on the number of parameters in thousands (K). For instance, U-Net has approximately 2 million parameters.}
}
\label{tab:results_tlc_bidepth}
\end{table}

The results highlight the effectiveness of integrating DeepShallow, ShallowDeep, and BiDepthNet encoder components with base TimeSeries Encoders like ConvLSTM and our ConvSelfAttention. Notably, enhancements with these architectures lead to significant improvements. For instance, when using ConvLSTM as the TimeSeries Encoder, its MSE decreases from $1.652$ to $1.515$ with the DeepShallow encoder, and to $1.500$ with the ShallowDeep encoder. The ConvLSTM combined with the full BiDepthNet encoder achieves an MSE of $1.479$. Notably, the full BDMNN (ConvSelfAttention + BiDepthNet) not only achieves the lowest MSE of $1.428$  but does so with a modest 440K parameters. This is significantly more efficient than several baselines like U-Net (2,000K) and CNN-RNN (2,400K), highlighting the effectiveness and parameter efficiency of our proposed architecture.

These improvements underscore the proposed method’s ability to more effectively exploit ST structures. By dynamically adjusting network depth via the BiDepth mechanism to balance short-term fluctuations and long-term trends, and by preserving spatial correlations through the ConvSelfAttention based TimeSeries Encoder, the BDMNN captures the inherent complexity of ST data better than baseline models and individual components alone. This synergy translates into more accurate predictions across diverse scenarios.

\paragraph{Spatial MAE Analysis}
\label{par:MARanalysis}

\begin{figure}[ht!]
\centering
\includegraphics[width=80mm]{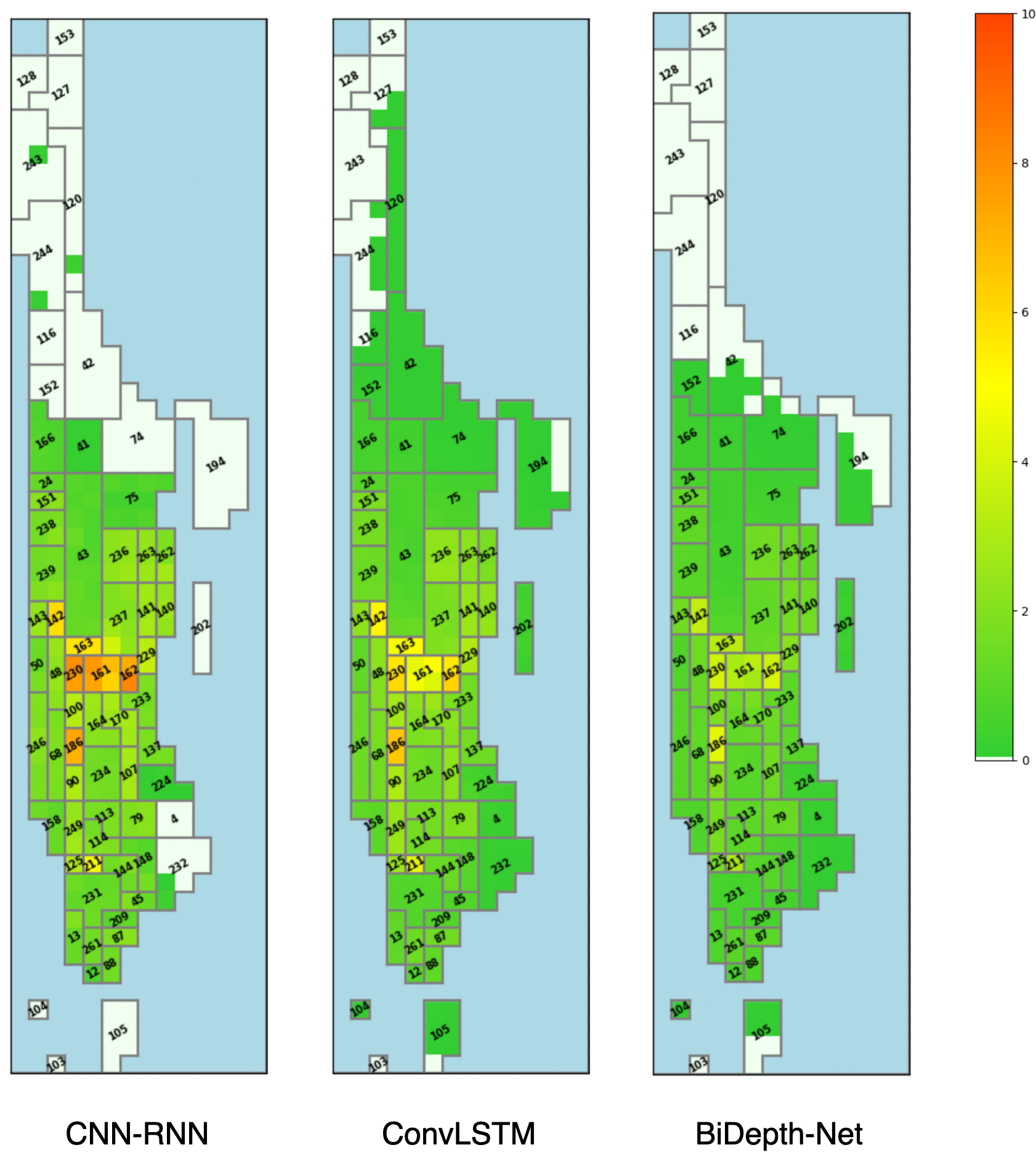}
\caption{Spatial MAE distribution over Manhattan for different models, based on the average MAE between prediction and the label across the whole test set and all time-stamps. The variance in heatmaps reveals prediction discrepancies across taxi zones. Distinctly, the BDMNN model (ConvSelfAttention + BiDepthNet) manifests consistently minimal errors throughout most zones, emphasizing its superior ST predictive capability.}
\label{fig:MAEMap}
\end{figure}

To better understand the spatial performance of the BDMNN relative to other benchmark models, we produced an MAE heatmap for the Manhattan region. This heatmap reflects the spatial distribution of prediction errors, averaged across the entire test set and all time-stamps, providing granular insight into where models succeed or struggle. As illustrated in \Cref{fig:MAEMap}, while each model exhibits areas of strength, the BDMNN consistently achieves notably lower error rates throughout Manhattan. Such a spatially resolved analysis demonstrates that the BDMNN not only excels in aggregate metrics like MSE but also delivers consistently lower per-zone errors. This spatial adaptability suggests that the BDMNN is better aligned with local ST dynamics, leading to more reliable predictions in diverse urban contexts.

\subsubsection{GPM Forecasting}

The results for the GPM dataset are presented in \Cref{tab:results_per_bidepth}. Consistent with the findings from the TLC dataset, the BiDepth variants outperformed the baseline TimeSeries Encoders. The full BDMNN (ConvSelfAttention + BiDepthNet) showed over a 14.58\% improvement when the BiDepthNet encoder was added to the ConvSelfAttention TimeSeries Encoder, and a ConvLSTM with BiDepthNet showed a 6.84\% improvement over the standalone ConvLSTM baseline. This consistent superiority across diverse datasets highlights the robustness and adaptability of the BDMNN architecture, indicating its potential for a wide range of ST forecasting applications. In the context of weather forecasting, improvements in MSE can yield substantial real-world benefits—enhancing resource planning, risk mitigation, and operational efficiency. The demonstrated improvements underscore BDMNN’s capacity to more effectively capture complex ST correlations, thus providing more reliable predictions in demanding scenarios.

\begin{table}[ht!]
\centering
\begin{tabular}{lccc}
    \toprule
    \textbf{Model} & \textbf{Model Complexity}\tnote{$\ast$}  & \textbf{MSE ($\pm$ Std. Dev.)} & \textbf{Improvements} (\%) \\
    \midrule
    ARIMA \citep{dubey2021study} & -   & $12.907 \pm 0.000$ & - \\
    SARIMA  \citep{dubey2021study} & - & $11.088 \pm 0.000$ & - \\
    U-Net \citep{ronneberger2015u} & 2,000 & $3.920 \pm 0.038$ & -   \\
    CNN-RNN  \citep{lu2020cnn} & 3,500 & $5.112 \pm 0.031$ & -  \\
    ViViT \citep{arnab2021vivit} & 4,200 & $3.112 \pm 0.021$ & - \\
    \hdashline
    ConvLSTM \citep{shi2015convolutional} & 1,520 & $3.071 \pm 0.011$ & -   \\
    + DeepShallow   & 1,600 & $2.779 \pm 0.018$ & +9.48 \\
    + ShallowDeep & 1,600 & $2.951 \pm 0.012$ & +3.87   \\
    + BiDepthNet & 1,680 & $2.861 \pm 0.019$ & +6.84 \\
    \hdashline
    ConvSelfAttention & 2,100 & $2.951 \pm 0.021$ & - \\
    + DeepShallow & 2,180 & $2.863 \pm 0.027$ & +2.95 \\
    + ShallowDeep & 2,180 & $2.861 \pm 0.023$ &  +3.02 \\
    + BiDepthNet & 2,260 & $\mathbf{2.520 \pm 0.019}$ &  +14.58 \\
  \bottomrule
\end{tabular}
\caption[Performance Comparison of BiDepth vs Benchmarks on GPM Data]{Performance comparison and parameter sizes of BiDepth versus Benchmarks on GPM Data. The model parameters are presented in thousands. The performance depicted is based on MSE loss (mean $\pm$ standard deviation from 5 runs). Models with modifications such as "+ DeepShallow", "+ ShallowDeep", and "+ BiDepthNet" indicate enhancements made to the base TimeSeries Encoders (ConvLSTM and ConvSelfAttention) using the BiDepth encoder components. The reported improvements reflect the percentage increase in performance (reduction in MSE) when incorporating these encoder components. \\
{ \footnotesize $\ast$ Model complexity is assessed based on the number of parameters in thousands (K). For instance, U-Net has approximately 2 million parameters.}
}
\label{tab:results_per_bidepth}
\end{table}


\subsection{Statistical Significance of Results}
\label{subsec:stat_sig}

To rigorously validate the performance gains offered by the BiDepth architecture, we conducted a statistical significance analysis using bootstrap resampling. For each dataset, we tested the null hypothesis that the mean MSE of the BiDepth-enhanced model is the same as its corresponding base model (ConvLSTM or ConvSelfAttention). The analysis was performed by generating 10,000 bootstrap samples for each comparison to compute empirical p-values.

The results, summarized in \autoref{tab:stat_sig}, show that the improvements achieved by BiDepth are statistically significant across all tested scenarios. In all four cases, the resulting p-value was less than 0.0001, indicating that the probability of observing such a large improvement in performance by random chance is exceedingly low. This allows us to reject the null hypothesis with high confidence and conclude that the BiDepth architecture provides a statistically significant improvement in predictive accuracy.

\begin{table}[ht!]
\centering
\caption{Statistical significance of MSE reduction by BiDepth enhancement.}
\label{tab:stat_sig}
\begin{tabular}{llcc}
\toprule
\textbf{Dataset} & \textbf{Comparison} & \textbf{Observed Mean MSE Reduction} & \textbf{p-value} \\
\midrule
NYC Taxi & BiDepth vs. ConvSelfAttention & 0.084 & \(< 0.0001\) \\
NYC Taxi & BiDepth vs. ConvLSTM          & 0.196 & \(< 0.0001\) \\
US Precipitation & BiDepth vs. ConvSelfAttention & 0.313 & \(< 0.0001\) \\
US Precipitation & BiDepth vs. ConvLSTM          & 0.204 & \(< 0.0001\) \\
\bottomrule
\end{tabular}
\end{table}

To further assess the consistency of this improvement, we conducted a direct pairwise comparison for each of the 100 experimental runs. This "win rate" analysis revealed that BiDepth achieved a lower MSE than its baseline counterpart in the vast majority of instances. Specifically, when enhancing ConvSelfAttention, BiDepth outperformed the baseline in 88\% of runs on the NYC Taxi dataset and 97\% of runs on the US Precipitation dataset. The advantage was even more pronounced with the ConvLSTM base, where BiDepth won in 99\% of runs for both datasets. 

This high degree of consistency, combined with the significant improvement in mean performance, robustly validates the architectural advantages of the BiDepth approach for spatio-temporal forecasting.

\subsection{Impact of Initial Depth in the BDMNN}
\label{sec:ImpactOfDepth}

The initial depth in our BDMNN, a significant hyperparameter, determines the extent of depth possibly needed to capture both historical and recent trends. To understand this relationship, we varied the initial depth and monitored its influence on the MSE in the Validation set as well as the training time for the TLC data.

\begin{figure}[ht!]
\centering
\includegraphics[width=120mm]{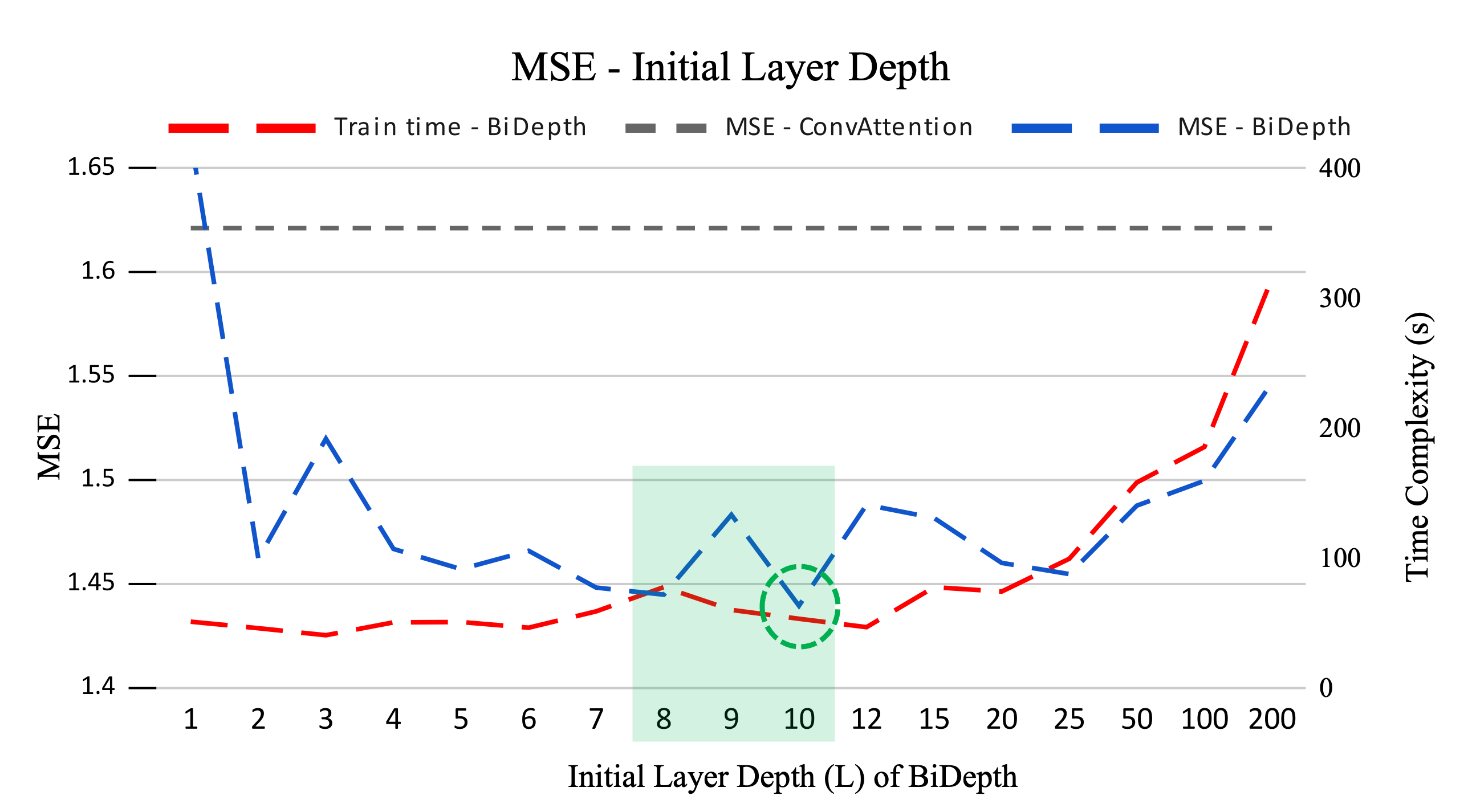}
\caption{A clear trend is observed, with MSE decreasing as depth increases up to a certain point, after which performance stabilizes while training time continues to rise. The left axis illustrates the MSE for both ConvAttention and BiDepth, while the right axis depicts the training time in seconds. The highlighted area and circled values are the recommended $L$ based on our analysis. This highlights the trade-off between enhanced performance and increased computational cost as model depth grows.}
\label{fig:MSEDepth}
\end{figure}

\Cref{fig:MSEDepth} demonstrates that as the initial depth amplifies, the MSE correspondingly drops, signaling improved model performance. However, there's a discernible inflection point beyond which additional depth no longer enhances accuracy, however it increases the training time. This optimal juncture provides a glimpse into the data's temporal nature, suggesting a balance between recent and historical data.

\section{Conclusion and Future Work}
\label{sec:conclusion}

In this paper, we introduced BDMNN, a novel deep learning architecture designed to discover and model the complex, multi-scale correlations inherent in ST data. BDMNN distinctively integrates two core innovations: a BiDepth mechanism that dynamically adjusts network depth to effectively balance long-term trends and short-term fluctuations, and a spatially-aware CSAC that preserves spatial integrity while capturing temporal dependencies. Our comprehensive evaluations on diverse real-world datasets, encompassing urban traffic and precipitation forecasting, demonstrated BDMNN's superior predictive accuracy and robustness compared to leading benchmark models. The results affirm the benefits of BDMNN's architectural design in capturing complex ST correlations and adapting to varied data dynamics.

While BDMNN shows significant promise, several avenues for future research could further enhance its capabilities and applicability. One potential direction is the exploration of more computationally efficient variants of the CSAC to improve scalability for extremely high-resolution ST datasets. Building on the current work, promising future research includes:
\begin{itemize}[leftmargin=*]
    \item Developing methodologies for an autonomous, data-driven configuration of the BiDepth mechanism's layer depths, moving beyond the fixed linear function used in this study to allow for more adaptive, sequence-specific architectures.
    \item Extending the BDMNN framework to incorporate diverse multi-modal inputs, such as exogenous variables or event data, to enrich contextual understanding for more nuanced predictions.
    \item Investigating the application of BDMNN in transfer learning scenarios, enabling knowledge leveraging across different ST forecasting tasks or geographical domains to improve generalization with limited data.
\end{itemize}
Pursuing these directions can further solidify BDMNN as a versatile and powerful tool for a wide range of ST prediction challenges.

\bibliographystyle{ACM-Reference-Format}
\bibliography{sample-base}










\end{document}